\documentclass[10pt,twocolumn,letterpaper]{article}

\usepackage{iccv}
\usepackage{times}
\usepackage{epsfig}
\usepackage{graphicx}
\usepackage{amsmath}
\usepackage{amssymb}

\usepackage{booktabs}
\usepackage{multirow}

\usepackage[pagebackref=true,breaklinks=true,letterpaper=true,colorlinks,bookmarks=false]{hyperref}

\iccvfinalcopy 

\def\iccvPaperID{24} 

\ificcvfinal\fi

\begin{document}

\title{MOFA: A Model Simplification Roadmap for Image Restoration on Mobile Devices}

\author{Xiangyu Chen\textsuperscript{\rm 1,2}$^\dagger$, Ruiwen Zhen\textsuperscript{\rm 2}$^\dagger$, Shuai Li\textsuperscript{\rm 2}, Xiaotian Li\textsuperscript{\rm 2}, Guanghui Wang\textsuperscript{\rm 3}$^*$\\
\textsuperscript{\rm 1}Department of EECS, University of Kansas, KS, USA \\ 
\textsuperscript{\rm 2}SenseBrain Technology, San Jose, CA, USA \\ 
\textsuperscript{\rm 3}Department of CS, Toronto Metropolitan University, Toronto, ON, Canada \\
\tt\small {xychen@ku.edu, \{zhenruiwen, shuai.li, lixiaotian\}@sensebrain.site,} \\ \tt\small {wangcs@torontomu.ca (* corresponding author)}
}

\maketitle
\def\thefootnote{$^\dagger$}\footnotetext{These authors contributed equally to this work}\def\thefootnote{\arabic{footnote}}
\ificcvfinal\thispagestyle{empty}\fi

\begin{abstract}

Image restoration aims to restore high-quality images from degraded counterparts and has seen significant advancements through deep learning techniques. The technique has been widely applied to mobile devices for tasks such as mobile photography. Given the resource limitations on mobile devices, such as memory constraints and runtime requirements, the efficiency of models during deployment becomes paramount. Nevertheless, most previous works have primarily concentrated on analyzing the efficiency of single modules and improving them individually. This paper examines the efficiency across different layers. We propose a roadmap that can be applied to further accelerate image restoration models prior to deployment while simultaneously increasing PSNR (Peak Signal-to-Noise Ratio) and SSIM (Structural Similarity Index). The roadmap first increases the model capacity by adding more parameters to partial convolutions on FLOPs non-sensitive layers. Then, it applies partial depthwise convolution coupled with decoupling upsampling/downsampling layers to accelerate the model speed. Extensive experiments demonstrate that our approach decreases runtime by up to 13\% and reduces the number of parameters by up to 23\%, while increasing PSNR and SSIM on several image restoration datasets. Source Code of our method is available at \href{https://github.com/xiangyu8/MOFA}{https://github.com/xiangyu8/MOFA}.
\end{abstract}

\section{Introduction}
During the past decade, deep learning has been the dominant approach for various computer vision tasks, including image classification \cite{ma2022semantic, yang2022unsupervised}, object detection \cite{li2021sgnet, xu2020adaptively}, segmentation \cite{rahman2023real, patel2022fuzzynet}, and image restoration   \cite{wang2022uformer,chen2021hinet,zamir2021multi,zamir2022restormer,dudhane2022burst,chu2022improving}, owing to its remarkable feature extraction capabilities derived from a substantial number of parameters. However, these large models often demand significant computational resources and result in longer inference latency, posing challenges for direct deployment on mobile devices. As a consequence, their application in mobile photography has been restricted.

To address this challenge, researchers have been exploring different approaches to simplify deep learning models by reducing the size of models directly, like channel dimensions and the number of layers    \cite{chen2022simple,lai2022face,wang2020practical}. Further, some works design efficient modules to replace inefficient modules    \cite{zamir2022learning,zhang2018shufflenet,ma2018shufflenet,howard2017mobilenets,sandler2018mobilenetv2,wang2023internimage,guo2023visual}, \eg replacing the original convolution with depthwise convolution    \cite{chen2022simple,wang2020practical}. However, most model compression methods focus on reducing the FLOPs or runtime only, at the cost of performance loss.

Partial Convolution (PConv)     \cite{chen2023run} is a recently proposed efficient module that simplifies original convolutions by performing calculations only on a portion of the channel dimension. In comparison to depthwise convolution, PConv demonstrates smaller runtime during deployment on mobile devices, even though it requires more parameters. Building on this observation, we can leverage the idea of adding more parameters during model compression to compensate for any potential performance loss, particularly in terms of PSNR (Peak Signal-to-Noise Ratio). By doing so, the model can still achieve faster runtime through model compression algorithms, while also benefiting from the larger capacity enabled by the additional parameters. As a result, the overall performance of the model is either maintained or even improved, considering the enhanced capacity brought about by the increased parameters.

Motivated by this concept, we propose a model simplification roadmap, which encompasses a set of tricks and techniques designed to optimize image restoration models for deployment on mobile devices. Through the application of these techniques, we aim to strike the right balance between model efficiency and performance, enabling the deployment of high-quality image restoration models on mobile devices without compromising on speed or output quality.  The contributions of this paper are summarized below.
\begin{itemize}
\item We propose a model simplification roadmap, consisting of tricks to add parameters to expand model capacity first and then reduce runtime, \ie \textit{\textbf{MO}re parameters and \textbf{FA}ster (MOFA)}. Given a small model, the model can be further accelerated before deployment with improved performance following the roadmap.
\item Building upon the concept of partial convolution, we propose Partial Depthwise Convolution (PDWConv) to further enhance acceleration. PDWConv involves performing depthwise convolution across only a portion of the channel dimension for input features.
\item We tested our roadmap on widely used small image restoration models, PMRID and NAFNet, on image restoration tasks including image denoising, image deblurring, and image deraining. Extensive experiments show that using our tricks, runtime for PMRID and NAFNet is decreased by up to 13\% and parameters are reduced by 23\% with better PSNR and SSIM.
\end{itemize}
\section{Related work}
This section presents related works on image restoration models on mobile devices as well as those modules designed to increase efficiency.
\begin{figure*}[h]
\centering
\includegraphics[width=0.71\linewidth]{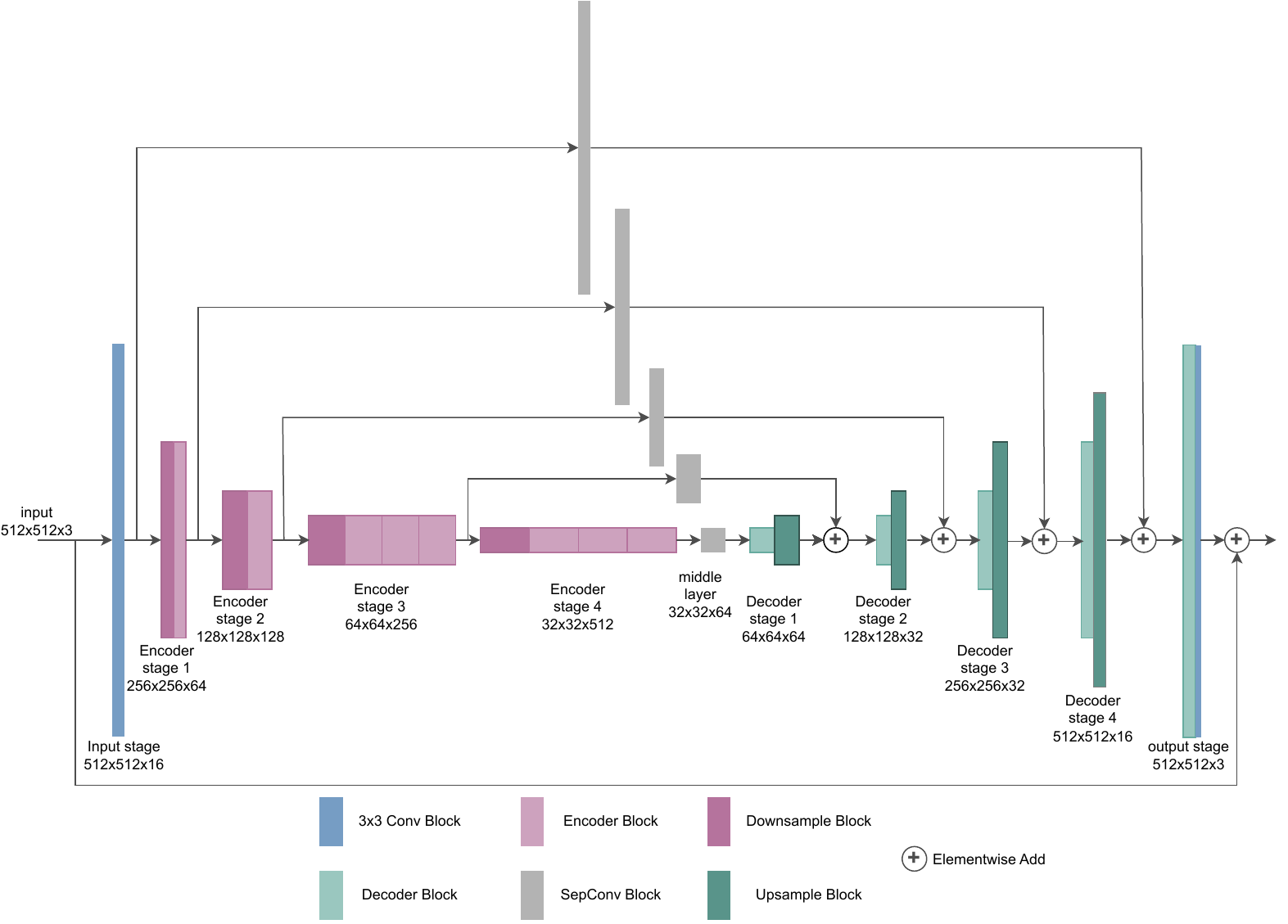}%
\caption{The overall network structure of PMRID}
\label{fig0:pmrid}
\end{figure*}
\begin{figure*}[h]
\centering
\includegraphics[width=0.71\linewidth]{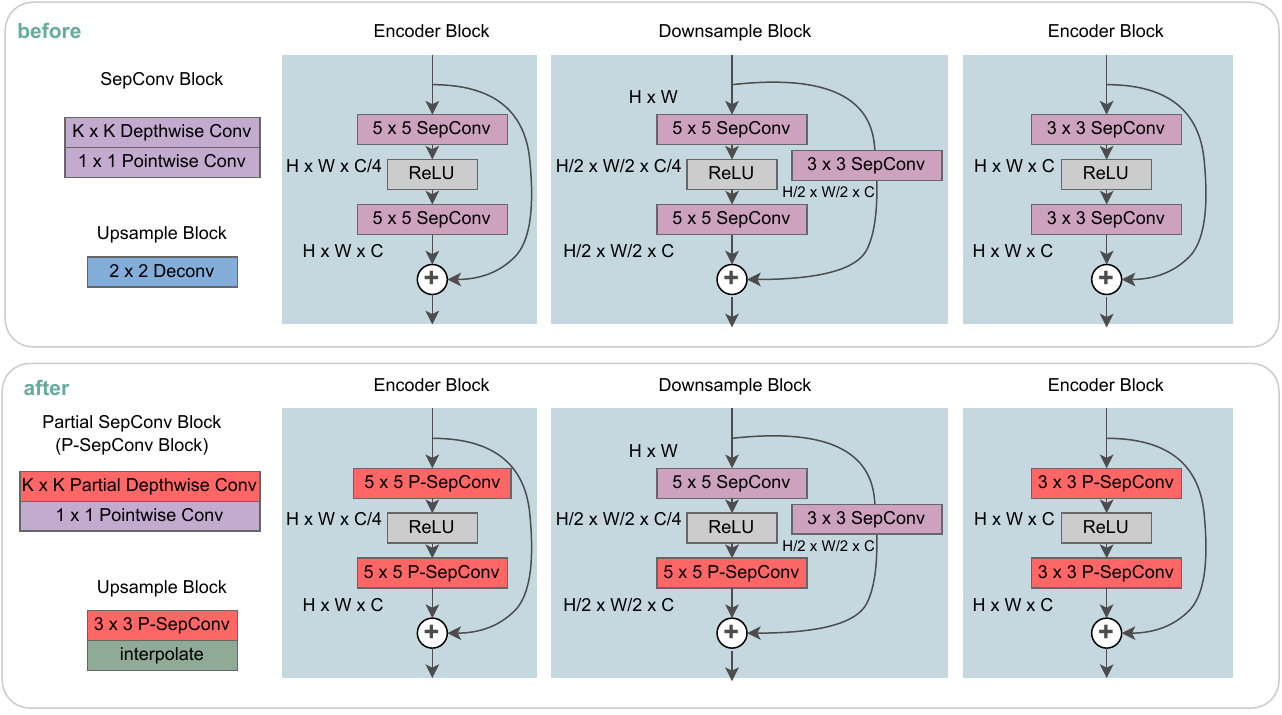}%
\caption{Detailed Blocks. We replace all Separable Convolutions (stride = 1) with one of the following convolutions: partial depth-wise convolution and partial convolution (PConv), with the portion 1/4 or 1/2.}
\label{fig0:blocks}
\end{figure*}
\subsection{Image restoration on mobile devices}
Image restoration aims to reverse degraded images into high-quality images. Tasks include image denoising, image deblurring, image draining, etc. These tasks are more commonly used in mobile devices like mobile photography. This encourages researchers to work on designing efficient deep learning models considering both runtime and image quality (IQ)     \cite{lahiri2020lightweight,guo2021fast,xu2021efficient,ignatov2022pynet,choi2023exploration,zhou2022thunder,yao2018fastdeepiot,justus2018predicting,choi2023ramit,chen2023vanillanet,zhang2023lightweight}. PMRID     \cite{wang2020practical} is one of the earliest efficient models. It carefully designs a U-Net-like model and increases efficiency by putting most parameters and calculations on the encoder stages. MicroISP     \cite{ignatov2022microisp} propose an attention-based multi-branch model so that it can be adapted to the different computational power of edge devices flexibly. MFDNet     \cite{liu2022mfdnet} designs a mobile-friendly attention module coupled with a reparameterization module to increase the denoising efficiency on iPhone 11. FusionNet     \cite{lai2022face} designs a U-Net-like model to take two images from dual cameras as input and output the deblurred images. Beyond image restoration, MobileOne     \cite{vasu2023mobileone} proposes re-parameterizable structures to decrease the runtime during inference for general tasks like image classification. Orthogonally, we propose a roadmap to further accelerate U-Net-like backbones.

\subsection{Efficient modules on computer vision}
The first set of efficient modules is related to the convolution operation. Vanilla convolution performs well but it visits each pixel the number of output channel times, which takes much computation and is inefficient for mobile applications. To improve this, Xception     \cite{chollet2017xception} proposes to use separable convolution, a depthwise convolution followed by a pointwise convolution, which requires visiting each pixel only once by depthwise convolution. This reduces computation greatly even counting the following pointwise convolution. PConv \cite{chen2023run} proposes to calculate partial convolution, \ie a portion like 1/4 of the input feature across channel dimension as they observe most kernels are learning similar features. 

In addition, some works propose a model simplification process by reducing unnecessary modules or replacing them with more efficient modules, \eg NAFNet     \cite{chen2022simple}, ConvneXt     \cite{liu2022convnet} and Efficient Unet     \cite{saharia2022photorealistic}. Other efficient modules include dynamic convolution     \cite{guo2023asconvsr}, fast nearest convolution     \cite{luo2022fast}, IRA block     \cite{conde2023perceptual}, and self-calibrated module     \cite{ma2022toward}.

However, all these consider only the efficiency of a single module and attempt to replace them with more efficient ones. Instead, we consider the efficiency of the entire network and manipulate inefficient layers according to the FLOPs distribution. PConv module provides us the flexibility to add parameters freely by setting the portion differently for different layers.

\section{Roadmap for simplification}

This section presents our model simplification roadmap, under the guideline that using more parameters wherever possible to compensate for the performance loss brought by rapid runtime decrease. We employ the widely used PMRID model from image denoising as an example for illustration. Following our roadmap, either the runtime gets reduced drastically or more parameters are added to the model in each step. At the end of optimization, the runtime is decreased by 11\%, the number of parameters is reduced by 6\% and the PSNR is increased by 0.02dB. Noticeably, this roadmap can be applied to any U-Net-like model as a model compression method before deployment on mobile devices. 

\textbf{Settings}
We train PMRID on SIDD dataset. Runtime is tested on one single v100 GPU with input size $3\times1024\times1024$, averaging 600 rounds. Each trick mentioned below from subsection \ref{trick1} is built on top of the previous ones.

\subsection{Starting point - PMRID}

\begin{figure}[t]
\centering
\includegraphics[width=0.95\linewidth]{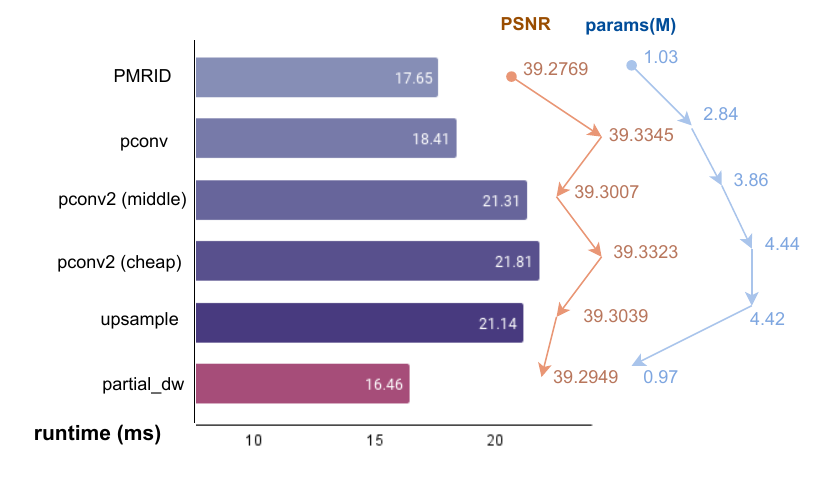}%
\caption{Roadmap. Starting from the baseline PMRID, we add each trick on top of all previous tricks, and it ends with partial\_dw.}
\label{fig1}
\end{figure}

PMRID     \cite{wang2020practical} is a U-Net-like model proposed for raw image denoising on mobile devices as in Figure \ref{fig0:pmrid} and \ref{fig0:blocks}. It is worth noting that PMRID holds much more parameters than normal U-Net-like image restoration models on mobile devices like DnCNN 
    \cite{zhang2017beyond} and NAFNet, but comparable runtime. This is because PMRID puts most of its parameters on its encoder only like half-UNet     \cite{lu2022half}, while most U-Net-like models distribute equally on the encoder and decoder. This asymmetrical design gives more freedom to put even more parameters on high-level encoders, where the input feature size is the smallest and hence the FLOPs will not increase drastically along with the increase of parameter amounts. For all convolutional layers, most layers use efficient depth-wise convolution followed by point-wise convolution, excluding the input and output layer, where vanilla convolutions are used as the channel dimension is relatively small hence it will not add too many calculations/FLOPs into the whole structure.

\subsection{Depthwise separable convolution to PConv} \label{trick1}
Partial convolution (PConv)     \cite{chen2023run} is a newly proposed efficient replacement for convolution modules on mobile devices, aiming to reduce both the number of floating-point operations (FLOPs), related to the number of multiply-adds, and the floating-point operations per second (FLOPS), considering the hardware deployment efficiency like memory access time. Unlike depthwise convolution which requires much memory access as convolution among different depths does not share parameters, partial convolution calculates only part of the convolution in terms of channel dimension, \eg $1/4$ by default. In this way, the computation gets reduced as depthwise convolution compared with the original convolution while it also enjoys the deployment time reduction due to parameter sharing as the original convolution, reducing memory access time drastically compared with depthwise convolution. Besides this, partial convolutions provide us more flexibility to manipulate later, \eg the portion. Hence, we first replace all depth-wise separable convolutions with partial convolutions. If there are vanilla convolutions exist (except the input and output layer), we also replace all those convolutions with partial convolutions directly. After this replacement, PSNR and SSIM increased to 39.3345 and 0.9559 respectively and runtime also increased by 0.76ms. This may be caused by the increase in parameters and FLOPs after the replacement. From now on, we use this partial convolution version as our model.

\subsection{More parameters - middle layers}
More parameters mean higher network capacity and better performance at most times. This motivates us to add more parameters to compensate for the PSNR and SSIM loss brought by the conversion from convolution to partial convolution. To achieve this, we can manipulate how large the ``part" is in partial convolution, \eg increasing it from default $1/4$ to $1/2$. Another problem is, where can we put those extra parameters to increase feature extraction ability without hurting runtime too much? 

As for image restoration tasks, they target on reconstruct pixels, not the semantic understanding of the whole images as high-level vision tasks, \eg image classification and detection. Thus we claim that more parameters should be added in shallow layers to increase low-level feature extraction power. However, more parameters in shallow layers usually mean much larger FLOPs and runtime as the feature size for shallow layers is usually larger than deeper layers. Considering both the performance and runtime, we increase the portion from $1/4$ to $1/2$ in partial convolution to add parameters on middle layers for both encoder and decoder, \eg, encoder 3 and decoder 2 for the PMRID model. As a result, PNSR and SSIM decreased a little bit, and runtime increased only 2.9ms even if the number of parameters increased from 2.84M to 3.86M and FLOPs from 2.12G to 3.25G. This increases model capacity. 

\subsection{More parameters - cheap layers}
\begin{figure*}[t]
\centering
\includegraphics[width=8.5cm,height=5 cm]{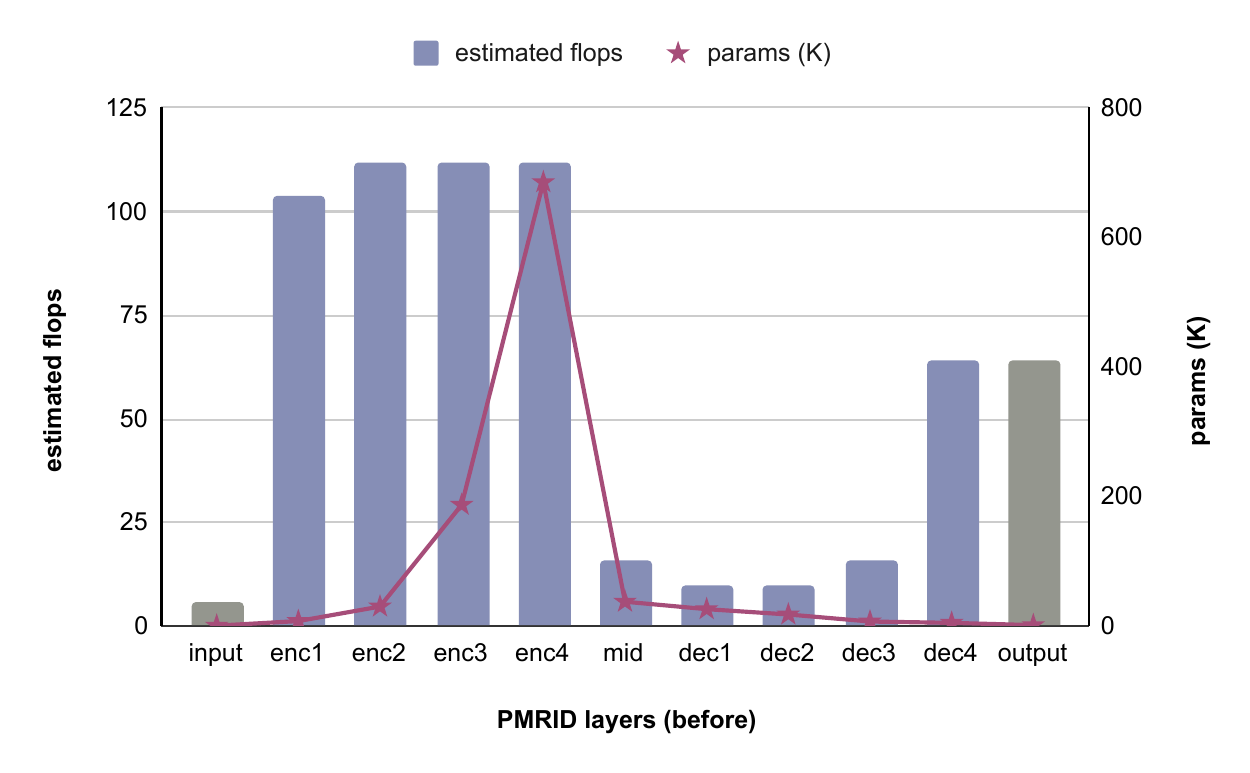}%
\includegraphics[width=8.5cm,height=5 cm]{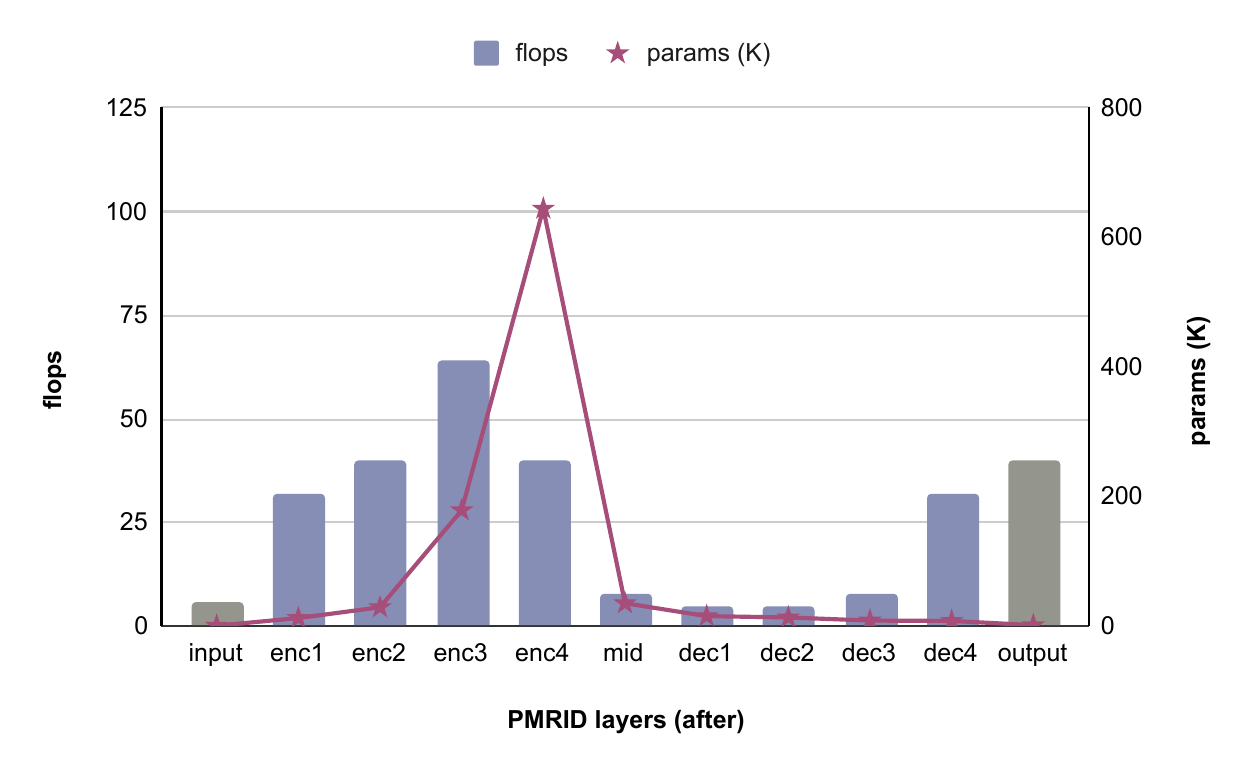}%

\caption{Estimated FLOPs distribution across layers before/after applying our roadmap tricks (dimension threshold $d*p$ is 0 here). Note that in PMRID structure, some layers (\eg input and output layers) are vanilla convolutions and separable convolutions elsewhere. To get a clear structure-related FLOPs distribution (only related to the channel dimensions and feature size changes), we estimate the FLOPs by assuming all layers are vanilla convolutions and excluding FLOPs in skip connections.}
\label{fig:dis}
\end{figure*}
To add more parameters without adding too many FLOPs and runtime, we first examine the flop distribution for each layer. There are two ways to achieve this. The most intuitive way is to calculate the FLOPs layer by layer using Python tools. However, this will include some unnecessary FLOPs we are not interested in here, \eg calculation in skip connections. Another way is to use a theoretical formula to estimate the FLOPs of convolution layers only. Here, we use the second method. The estimated FLOPs distribution can be found in Figure \ref{fig:dis}. It shows that all the decoders along with the middle layer take up relatively fewer FLOPs compared with encoders, hence we call them ``cheap layers". As a result, we increase the portion from $1/4$ to $1/2$ for partial convolutions in decoder layers. After adding this trick, PSNR, and SIMM increased to 39.3323 and 0.9560 respectively while runtime only increased 0.5ms. The amount of parameters now is 4.44M, which is 4$\times$ compared with the original PMRID model. After adding parameters to increase the model capacity using previous tricks, the following subsections focus on reducing the runtime. 
\subsection{Faster - upsampling/downsampling}

Downsampling and upsampling modules are vital in U-Net-like networks. There are some choices for them, \eg PMRID used depthwise separable convolution with stride 2 for downsampling and deconvolution for upsampling while NAFNet employed convolution coupled with pixel shuffle for upsampling. Instead, paper     \cite{saharia2022photorealistic} argued that performing downsampling before the convolution and upsampling after the convolution can improve the efficiency, different from the normal order for downsampling and upsampling. Hence, we may first decouple downsampling from convolution with stride 2 to one average pooling layer followed by a convolution layer with stride 1. Upsampling deconvolutions are replaced with an interpolation layer following a stride-1 convolution layer. We select the most efficient downsampling/upsampling from the choices mentioned above and the original downsampling/upsampling strategies for different structures. For PMRID, we replace only the upsampling layers with the decoupled ones. After this change, the runtime gets reduced by 0.67 ms, PSNR dropped a little bit while SSIM remains the same.

\subsection{Faster - PConv to PDWConv}
\begin{figure}[t]
\centering
\includegraphics[width=8cm,height=5.5 cm]{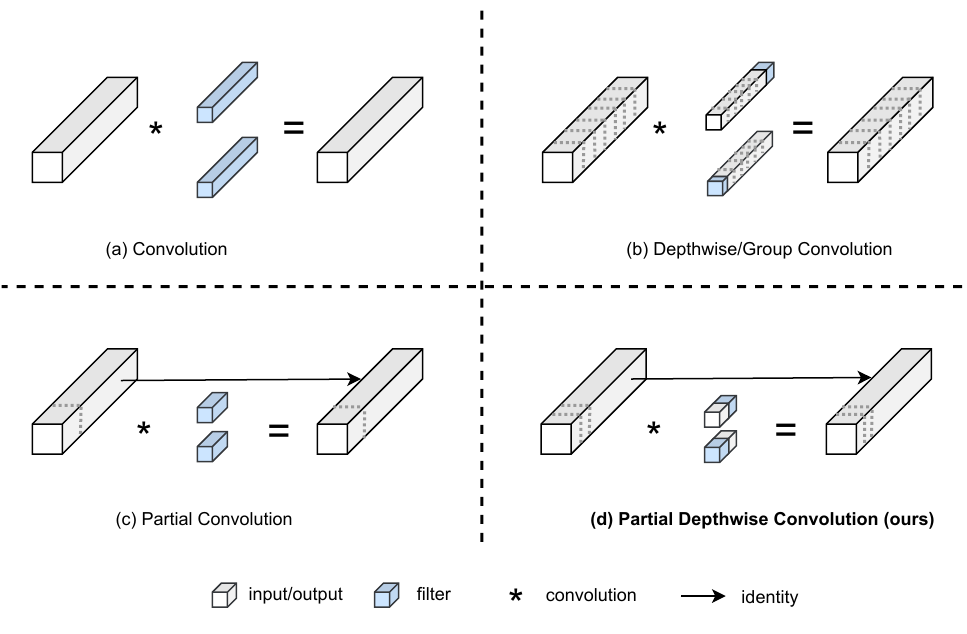}%
\caption{The structure of vanilla convolution, depthwise convolution, partial convolution, and partial depthwise convolution.}
\label{fig2:pdconv}
\end{figure}

To further accelerate it, we propose to replace partial convolution (PConv) with partial depthwise convolution (PDWConv). As illustrated in Figure \ref{fig2:pdconv}, Partial depthwise convolution calculates only a portion across the channel dimension of the input features as in PConv and the rest untouched part uses identity mapping. Unlike PConv, we perform depthwise convolution for the selected portion while PConv uses original convolution. 

Besides this, noticing that after taking the portion across the channel dimension, the input feature dimension will be decreased drastically. For features with small channel dimensions, \eg 64 or 32, only 16 or 8 channels are used to learn the parameters, which may hurt the performance worse than larger features. Thus, we replace only those PConv with PDWConv for larger channel dimensions while keeping PConv for smaller ones. For the PMRID model, we set the dimension threshold $d*p$ ($d$ is the input channel dimension for the current layer and $p$ is the portion) to 32 according to the experimental results. 

Finally, the runtime gets decreased to $16.46$ms while PSNR and SSIM become 39.2769 and 0.9556 respectively, still better than the original PMRID backbone.

\section{Experiments}
This section presents a detailed ablation study and analysis of some tricks in enhancing PMRID model on SIDD denoising dataset and both quantitative and qualitative results when we accelerate other models on RGB image denoising, image deblurring, raw image denoising, image deblurring with JPEG artifacts and image deraining tasks.
\subsection{Setting}
To verify the effectiveness of our pipeline, we apply it on two different models, NAFNet-tiny     \cite{chen2022simple} which has only 7 blocks, and the backbone from PMRID     \cite{wang2020practical}. For both models, our settings are mainly based on NAFNet     \cite{chen2022simple} as PMRID lacks the settings and reported results on most image restoration benchmarks. Adam optimizer     \cite{kingma2014adam} is used with $\beta_1 = 0.9$, $\beta_2 = 0.9$ and weight decay 0. Total iteration is $200K$ during training. In addition, we report the FLOPs (input size $3\times256\times256$) and the number of parameters for comparison. For the performance, we use PSNR and SSIM for quantitative comparison. Qualitative visualization can be found in Section \ref{sec:vis}. We tested runtime using two ways. The first one is gpu based as mentioned above. We tested on a single v100 with $3\times1024\times1024$ as the input size, averaging from 600 rounds. 
 Besides this, we also test our models on Android phone Tecno Camon 19 Pro based on Pytorch Mobile framework, using image size $3\times256\times256$,  averaging runtime from 3 forward passes. We use a larger input size on GPU runtime testing because when both our model and input are too small, the variance of runtime would be comparable to the actual runtime on v100 GPU, leading to biased averaged runtime.

\begin{table*}[h]
  \begin{center}
\begin{tabular}{c|ccc|cc||cc|cc|cc}
\toprule
& \multirow{2}{*}{PConv} & PConv2 & PConv2 & \multirow{2}{*}{down/up} & \multirow{2}{*}{PDWConv}&FLOPs & Param & \multirow{2}{*}{PSNR($\uparrow$)} & \multirow{2}{*}{SSIM($\uparrow$)} & \multicolumn{2}{c}{runtime (ms) ($\downarrow$)}\\ 
& &(middle) &(cheap) & & &(G)& (M)  & &&GPU & Mobile\\ \hline
baseline &  &&& & & 1.11&1.03& 39.2769 & 0.9556 & 17.65 & 220\\ \hline
  & \checkmark &  &&& &2.12&2.84& \textbf{39.3345}& \textbf{0.9559}& 18.41&\textbf{212}\\ \hline
    & \checkmark& \checkmark & &&&3.25&3.86& \textbf{39.3007}&\textbf{0.9557}& 21.31 & 250 \\
     & \checkmark  & & \checkmark & &&2.69&3.46&\textbf{39.3211}&\textbf{0.9559}&19.19 & 218\\
     & \checkmark  &\checkmark & \checkmark & &&3.69&4.44&\textbf{39.3323}&\textbf{0.9560}&21.81 & 255\\ \hline

      & & & & \checkmark &&0.95&1.01 &39.2741&0.9556 & \textbf{16.99} & \textbf{153} \\ \hline
       & \checkmark & & &  & \checkmark& 1.03&0.95& 39.2104 & 0.9552 & \textbf{14.46} & \textbf{162}\\ \hline
       &\checkmark &\checkmark& \checkmark & \checkmark& \checkmark & 1.11&0.97&\textbf{39.2949}& \textbf{0.9558} &  \textbf{16.46} & \textbf{196}\\
\bottomrule
\end{tabular}
\end{center}
\caption{Ablation study based on PMRID model for image denoising on SIDD dataset.}
\label{table:ab}
\end{table*}

\subsection{Ablation study}
In this subsection, we examine how each trick works in this acceleration process. Results can be found in Table \ref{table:ab}. We can find the first 3 tricks mainly increased the PSNR and SSIM by adding more parameters efficiently without adding too much runtime. Specifically, replacing with partial convolution increased PSNR and SSIM to 39.3345 and 0.9559 respectively compared with the baseline (PSNR: 39.2769, SSIM:  0.9556). Adding more parameters for middle layers also boosts PSNR and SSIM to 39.3007 and 0.9557 separately. Adding more parameters on cheap layers coupled with middle layers together further increased PSNR to 39.3323 and SSIM to 0.9560. All these tricks bring much more parameters to the model to increase the model capacity without introducing too many FLOPs and runtime, making space for the runtime reduction tricks to keep PSNR and SSIM. 

For the upsampling/downsampling modules, PMRID uses deconvolution layers to perform upsampling. When we decouple it into a convolution layer and an upsampling layer, the runtime gets reduced by 0.65ms while PSNR and SSIM are still comparable.

Finally, we replace all PConv with PDWConv, \ie we use depth-wise convolution to replace the original convolution in PConv and it results in partial depth-wise convolution as illustrated in Figure \ref{fig2:pdconv}. When we apply this directly on PConv skipping all steps in the middle, performance dropped from (PSNR: 39.3345, SSIM: 0.9559) to (PSNR: 39.2104, SSIM: 0.9552), even if runtime dropped significantly. This also demonstrates the necessity of expanding the model capacity to increase the model performance in the middle steps.

\subsection{Analysis of different components}
This subsection presents experimental results of how to determine alternatives for each trick.

\begin{table}[h]
  \begin{center}
\begin{tabular}{l|cc|cc}
\toprule
layers & FLOPs (G) & Param(M) & PSNR & SSIM \\ \hline
baseline & 1.11 & 1.03 & 39.2769 & 0.9556 \\ 
+PConv & 2.12 & 2.84 & 39.3345 & 0.9559 \\ \hline
enc3, dec2 & 3.25 & 3.86 & 39.3007 & 0.9557\\
enc2, dec3 & 2.58 & 2.93 & 39.2508 & 0.9556 \\
enc1, dec4 & 2.54 & 2.86 & 39.2770 & 0.9556 \\
\bottomrule
\end{tabular}
\end{center}
\caption{Adding more parameters to middle layers on PMRID.}
\label{table:mid}
\end{table}

\textbf{Analysis of middle layers} As mentioned above, we aim to add more parameters in middle-level layers to balance between performance and runtime, as we believe more parameters in shallower layers would be more appealing for low-level vision tasks as they focus more on extracting low-level features. However, more parameters in shallow layers also mean much more FLOPs as they take the largest feature size in the entire U-Net-like network. As a result, we decide to add parameters on the middle layers by increasing the portion of PConv from 1/4 to 1/2. The experiments can be found in Table \ref{table:mid}. 

As adding more parameters on PConv layers in encoder stage 3 and decoder stage 2 increased PSNR from 39.2769 to 39.3007 compared with the PMRID backbone baseline while other encoder-decoder pairs even dropped the PSNR and SSIM, we use encoder stage 3 and decoder stage 2 as our final middle layers to add parameter in PMRID. Besides this, this pair also introduces much more parameters than others. This is because encoder stage 3 in this pair has 4 encoder blocks while other encoder stages have only 2.

\begin{figure*}[ht]
\centering
\includegraphics[width=0.9\linewidth]{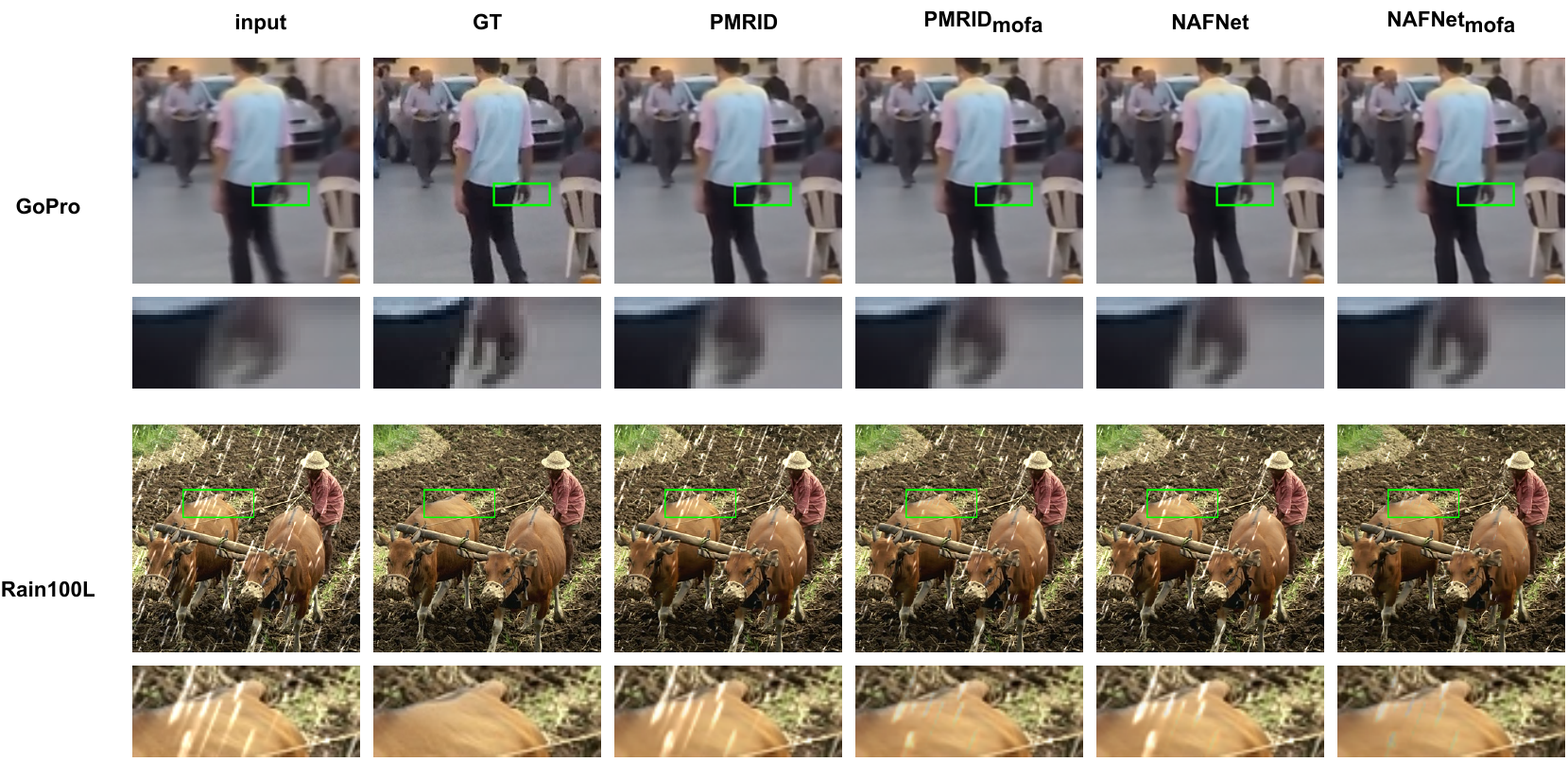}%
\caption{Visualization. Image restoration results from PMRID, PMRID$_{mofa}$, NAFNet and NAFNet$_{mofa}$ on different tasks.}
\label{fig3:vis}
\end{figure*}

\begin{table}[h]
  \begin{center}
\begin{tabular}{l|cc|cc}
\toprule
replacement  & Param(M) & runtime& PSNR & SSIM \\ \hline
no &  4.44 & 255 & 39.3323 & 0.9560 \\ 
upsampling &  4.42 & 268& 39.3039 & 0.9561\\
downsampling &  4.61 & 460& 39.3334 & 0.9559 \\
\bottomrule
\end{tabular}
\end{center}
\caption{Replacement of upsampling/downsampling in PMRID.}
\label{table:up}
\end{table}

\begin{table*}[h]
  \begin{center}
\begin{tabular}{l|llc|cc|cc|cc}
\toprule
\multirow{2}{*}{Method}&runtime (ms) &\multirow{2}{*}{\# params} &FLOPs&\multicolumn{2}{c|}{denoising (SIDD)} &\multicolumn{2}{c|}{debluring (GoPro)} & \multicolumn{2}{c}{debluring (REDS)}\\
 & Mobile&  &(G)& PSNR & SSIM & PSNR & SSIM & PSNR & SSIM \\ \hline
PMRID & 220$^*$&1.03M&1.15 &39.2769& 0.9556 &29.0840&0.9153 &27.6501& 0.8348\\ 
PMRID$_{mofa}$ &196(\footnotesize{\textit{-11\%}})&0.97M(\footnotesize{\textit{-6\%}}) &1.11& \textbf{39.2949} & \textbf{0.9558} &\textbf{29.2274} & \textbf{0.9176} & 27.6493 & 0.8347\\
\hline
NAFNet  & 69$^*$&286.6K&1.03 &39.2042&0.9555 &29.4337&0.9237&27.6747&0.8381\\
NAFNet$_{mofa}$  &60(\footnotesize{\textit{-13\%}}) &222.1K(\footnotesize{\textit{-23\%}})&0.94 &\textbf{39.3098}&\textbf{0.9560} & \textbf{29.4810} & \textbf{0.9242} & \textbf{27.7450} & \textbf{0.8398}\\
\bottomrule
\end{tabular}
\end{center}
\caption{Image denoising and deblurring results on different datasets. $^*$ Represents we also added \textit{split\_cat} operations for fair comparison.}
\label{table:denoise}
\end{table*}

\begin{table*}[h]
  \begin{center}
\begin{tabular}{l|cc|cc|cc|cc|cc}
\toprule
\multirow{2}{*}{Method}&\multicolumn{2}{c|}{Test100}&\multicolumn{2}{c|}{Rain100H} & \multicolumn{2}{c|}{Rain100L}&\multicolumn{2}{c|}{Test2800} &\multicolumn{2}{c}{Test1200} \\
&PSNR&SSIM  &PSNR & SSIM& PSNR & SSIM & PSNR & SSIM & PSNR & SSIM \\ \hline
PMRID &28.3400 &0.8690&27.9492&0.8343&32.5887&0.9316&32.6028 &0.9260&31.4316 &0.9001\\ 
PMRID$_{mofa}$ &28.2158 &\textbf{0.8739} &\textbf{28.3177} & \textbf{0.8453} &\textbf{32.7912} & \textbf{0.9330}&\textbf{32.6106} &\textbf{0.9272}& \textbf{31.4936} & \textbf{0.9034}\\
\hline
NAFNet &28.0719&0.8775&27.8863&0.8414&33.6056&0.9467&32.8942&0.9298 &32.6602&0.9181\\
NAFNet$_{mofa}$ &\textbf{28.9186} & \textbf{0.8888}& \textbf{28.5133} & \textbf{0.8531}& \textbf{33.9560} & \textbf{0.9500} &\textbf{32.9450} &\textbf{0.9303} & 32.4535 & 0.9150\\

\bottomrule
\end{tabular}
\end{center}
\caption{Image deraining results on different test datasets.}
\label{table:derain}
\end{table*}

\textbf{Analysis of upsampling/downsampling layers} For upsampling, we replace deconvolution layers with convolution and an upsampling layer in PMRID. This order can decrease the FLOPs compared with the reverse. For downsampling, the downsampling layer coupled with the convolution layer (stride 1) replaced the stride 2 convolution layers. Results can be found in Table \ref{table:up}. Here, the runtime is based on Pytorch Mobile tested on Android phones. We can find adding upsampling can reduce GPU runtime a little bit as in Table \ref{table:ab} while increasing inference time on the device as in Table \ref{table:up}. This may be caused by different acceleration implementations for GPU and Pytorch Mobile on devices. Besides this, downsampling increased runtime on Pytorch Mobile a lot for PMRID, from 255ms to 460ms. Hence, we replace upsampling only for PMRID.

\begin{table}[h]
  \begin{center}
\begin{tabular}{c|cc|cc}
\toprule
thres $d*p$ & runtime(ms) & Param(M) & PSNR & SSIM \\ \hline
0 & 14.10 & 0.94 & 39.2513 & 0.9554 \\
 16& 15.8& 0.95 & 39.2408 & 0.9557 \\ 
32 & 16.46 & 0.97 &39.2949&0.9558 \\
64&16.56 & 1.21 & 39.2920 & 0.9559 \\
128 &16.71 & 1.25 & 39.2973 & 0.9561 \\
\bottomrule
\end{tabular}
\end{center}
\caption{The effect of threshold $d*p$ to replace PConv with PDWConv.}
\label{table:thres}
\end{table}

\textbf{Analysis of PDWConv threshold} Noticing that PDWConv calculates only a portion of the entire depthwise convolution across channel dimension, and when the dimension is relatively small, \eg 32, it may hurt performance drastically compared with the convolution in PConv which hold much more parameters. As a result, we replace only PConv with PDWConv with a larger channel dimension and set a threshold to determine. If the channel dimension of input features $d$ divided by the portion $p$, \ie $d*p, p = [1/2,1/4]$, the actual channel dimension to be calculated, is larger than the threshold, replace PConv with PDWConv, otherwise, we keep them PConv. Considering both runtime and PSNR/SSIM, we choose 32 as the threshold for PMRID. It may differ for different backbones.

\subsection{Applications}
We apply our roadmap with two different backbones, PMRID     \cite{wang2020practical} and NAFNet     \cite{chen2022simple}, and get their accelerated versions PMRID$_{mofa}$ and NAFNet$_{mofa}$. As layer normalization is not deployable on mobile devices, we remove this layer directly to test runtime. We tested these models on various image restoration tasks, including denoising, deblurring, and deraining.

\textbf{RGB Image Denoising.} We test denoising performance on SIDD dataset. Quantitative results are shown in Table \ref{table:denoise}. This table shows for both PMRID and NAFNet, our accelerated versions decreased runtime on mobile devices by $11\%$ and $13\%$, and PSNR and SSIM are also increased at the same time, from 39.2769dB to 39.2949dB for PMRID and from 39.2319dB to 39.3098dB for NAFNet. Note that runtime reduction is applied to all the following tasks as we use the same models for comparison.

\textbf{Image Deblurring.} We tested image deblurring performance on GoPro     \cite{nah2017deep} dataset. As shown in Table \ref{table:denoise}, we have increased PSNR by 0.14dB for PMRID and 0.05dB for NAFNet while reducing the runtime by more than $10\%$ for both backbones.

\textbf{Image Deblurring with JPEG artifacts.} As in NAFNet     \cite{wang2020practical}, we also tested the image deblurring effect with JPEG artifacts. Following settings in NAFNet     \cite{wang2020practical}, we use REDS     \cite{nah2021ntire} dataset and evaluate on REDS-val-300. Table \ref{table:denoise} shows NAFNet$_{mofa}$ increased PSNR by 0.07dB compared with NAFNet while the runtime gets decreased and PMRID$_{mofa}$ is comparable to PMRID.

\textbf{Image deraining.} Following settings in HINet     \cite{chen2021hinet}, we also train our models on image deraining task with Rain13k dataset and test on Test100     \cite{zhang2019image}, Rain100H     \cite{yang2017deep}, Rain100L     \cite{yang2017deep}, Test2800     \cite{fu2017removing} and Test1200     \cite{zhang2018density} datasets. Quantitative results are shown in Table \ref{table:derain}. From this table, our accelerated versions for both backbones have improved the performance for most deraining datasets while reducing the runtime. Specifically, PMRID$_{mofa}$ defeats PMRID on Rain100H, Rain100L, Test2800 and Test1200 datasets, and NAFNet$_{mofa}$ improves over NAFNet on Test100, Rain100H, Rain100L and Test2800 datasets.

\subsection{Visualization}\label{sec:vis}
To understand the qualitative results, we visualize image restoration results on different datasets, SIDD dataset for image denoising, GoPro dataset for image deblurring, REDS dataset for image deblurring with JPEG artifacts, and Rain100L dataset for deraining as in Figure \ref{fig3:vis}.

For image denoising, we can find all these models, PMRID, PMRID$_{mofa}$, NAFNet and NAFNet$_{mofa}$ have good denoising ability. Among them, we observe NAFNet$_{mofa}$ is the best.

In GoPro dataset, the quality of deblurring results increases from PMRID to PMRID$_{mofa}$ where the shape of the hand in the figure changes from blurred for PMRID to clear for NAFNet$_{mofa}$. 

For deraining results on Rain100L dataset, we can see both PMRID$_{mofa}$ and NAFNet$_{mofa}$ show better deraining results than their baselines.

\section{Conclusion and limitation}


This paper has presented a comprehensive roadmap to accelerate image restoration models before deploying them on mobile devices. The key idea is to enhance the model's capacity without significantly increasing runtime by strategically adding more parameters to FLOP-insensitive layers and middle layers. Additionally, we propose the decoupling of upsampling/downsampling layers and apply partial depthwise convolution to further accelerate the models. The proposed roadmap is effectively applied to PMRID and NAFNet for various image restoration tasks. Extensive experimental results demonstrate the effectiveness of our strategy. Following our approach, we achieve a runtime reduction of up to 13\% and reduce the number of parameters by 23\% while simultaneously improving the image restoration performance. 

Our approach offers a powerful solution for deploying image restoration models on mobile devices, striking a balance between increased model capacity and faster inference without compromising restoration quality.
For deployment, the decrease of parameters is suitable for parameters-sensitive implementation like sensor programming. The introduction of $split\_cat$ operation in PDWConv and PConv might be less efficient in deployment based on Pytorch Mobile, ONNX and Caffe and need to be further accelerated.

\section*{Acknowledgement}
The work was supported by and conducted at SenseBrain Technology. G. Wang was partly supported by the Natural Sciences and
Engineering Research Council of Canada (NSERC).
{\small
\bibliographystyle{ieee_fullname}
\bibliography{egbib}
}

\end{document}